\documentclass[letterpaper]{article}
\usepackage{aaai}
\usepackage{times}
\usepackage{helvet}
\usepackage{courier}
\usepackage[pdftex]{graphicx}
\usepackage[]{amsmath, mathtools,}
\usepackage{algpseudocode}
\usepackage{algorithm}
\usepackage{placeins}
\begin{document}
\title{Predicting Food Security Outcomes Using CNNs for Satellite Tasking}
\author{Swetava Ganguli, Jared Dunnmon, Darren Hau\\
Department of Computer Science, 353 Serra Mall, Stanford, CA 94305\\
{\{swetava, jdunnmon, dhau\}@cs.stanford.edu}
}
\maketitle
\begin{abstract}
\begin{quote}
Obtaining reliable data describing local Food Security Metrics (FSM) at a granularity that is informative to policy-makers requires expensive and logistically difficult surveys, particularly in the developing world. We train a CNN on publicly available satellite data describing land cover classification and use both transfer learning and direct training to build a model for FSM prediction purely from satellite imagery data. We then propose efficient tasking algorithms for high resolution satellite assets via transfer learning, Markovian search algorithms, and Bayesian networks.
\end{quote}
\end{abstract}

\section{Introduction and Background}
In a recent paper (Neal Jean 2016), Convolutional Neural Networks (CNNs) were trained to identify image features that explain up to 75\% of the variation in local level economic outcomes. This ability potentially enables policy-makers to measure need and impact without requiring the use of expensive and logistically difficult local surveys (\cite{un14}, \cite{un15}, \cite{devarajan13} and \cite{jerven13}). This ability is promising for some African countries, where data gaps may be exacerbated by poor infrastructure and inadequate government resourcing. According to World Bank data, during the years 2000 to 2010, 39 of 59 African countries conducted fewer than two surveys from which nationally representative poverty measures could be constructed. Of these countries, 14 conducted no such surveys during this period (World Bank 2015). Closing these data gaps with more frequent household surveys is likely to be both prohibitively costly, perhaps costing hundreds of billions of U.S. dollars to measure every target of the United Nations Sustainable Development Goals in every country over a 15-year period (\cite{jervenbc14}) and institutionally difficult, as some governments see little benefit in having their lackluster performance documented (\cite{devarajan13}, \cite{sg15}). A popular recent approach leverages passively collected data like satellite images of luminosity at night (“nightlights”) to estimate economic activity (\cite{henderson_2011}, \cite{chen_2011}, \cite{pinkovskiy_2017}).

Over the last several years, an increasing supply of satellite data has enabled a variety of important studies in areas of environmental health, economics, and others \cite{planet2015pres}.  The unique insight attainable from such datasets has continued to be integrated into the intelligence operations of both public and private sector entities to the point that a number of modern processes would not be possible in their absence \cite{meyer2014spysats}. Additionally, the increasingly common integration of hyperspectral sensors into commercial satellite payloads has extended the types of conclusions that can be drawn from satellite imagery alone \cite{romero2014unsupervised}. Recent developments deep CNNs have greatly advanced the performance of visual recognition systems. In addition, using convolutional neural networks and GANs on geospatial data in unsupervised or semi-supervised settings has also been of interest recently; especially in domains such as food security, cybersecurity, satellite tasking, etc. (\cite{ganguli2019geogan}, \cite{dunnmon2019predicting}, \cite{perez2019semi}). In this paper, we demonstrate the use of CNNs for predicting food security metrics in poor countries in Africa using a combination of publicly available satellite imagery data from the work of \cite{basu15} (``DeepSat'' dataset) and UN data made available by the Stanford Sustainability Lab (``SustLab'' dataset).\footnote{Note that the DeepSat dataset is available online while the SustLab dataset is currently proprietary.} We also demonstrate the use of these prediction models for tasking of relatively expensive, high-resolution satellites in a manner that maximizes the amount of imagery collected over food insecure areas. The task considered in this paper is that of reaching the region of lowest food security while traversing a path that is globally optimal in the sense that the most food-insecure regions possible are traversed. 

Our project thus consists of two main challenges: (i) designing a CNN that ingests satellite images as an input and outputs successful predictions of key food security metrics such as z-scores of stunting or wasting, and (ii) using the food security predictions as rewards in a reinforcement learning process to train a satellite to automatically traverse areas of interest. The first task is a machine-learning-intensive task which requires using a large dataset of images to train a CNN model. We first utilize a publicly available labeled dataset of 28 x 28 image tiles at 1m resolution labeled with one of six ground cover classifications to train a CNN focused on making predictions from overhead imagery \cite{basu15}. This trained network can be used to extract features from satellite imagery, which can in turn be used as the input to a ridge regression procedure aimed at accurately predicting food security metrics. Alternatively, we can also use both direct learning and transfer learning to enable direct prediction of food security metrics from a pure CNN model, which will represent an interesting comparison to the regression procedure. We choose to use convolutional neural networks for this task due to their superior performance on image classification problems. Their performance generally results from effective integration of spatial locality in convolutional features that aids accurate classification predictions \cite{hinton12}. We use a relatively simple VGG-style architecture in this work \cite{zisserman14}. 

The second task is using the output FSMs to define the state space for a set of path search algorithms. The general task of training a model to change the orientation of a satellite can be modeled as the problem of an agent finding its way to a high terminal reward on a grid via a Markovian optimal path. For brevity, we will chose demonstrative numbers for easier explanation. Consider a very large image (e.g. 400 x 400 pixel image showing a 1000 X 1000 km area). Let us call our trained CNN H. We train the CNN on a dataset of 28 x 28 images (corresponding to 28 m x 28 m areas) and regress on a dataset of values of a given poverty/food security metric so that when an image is provided to the trained network, the output is a number quantifying this metric. More precisely: \\

28 x 28 Resolution Image $\implies$ H[Image] $\implies$ A number corresponding to food security metric.\\

\begin{figure}
\includegraphics[width=0.5\textwidth]{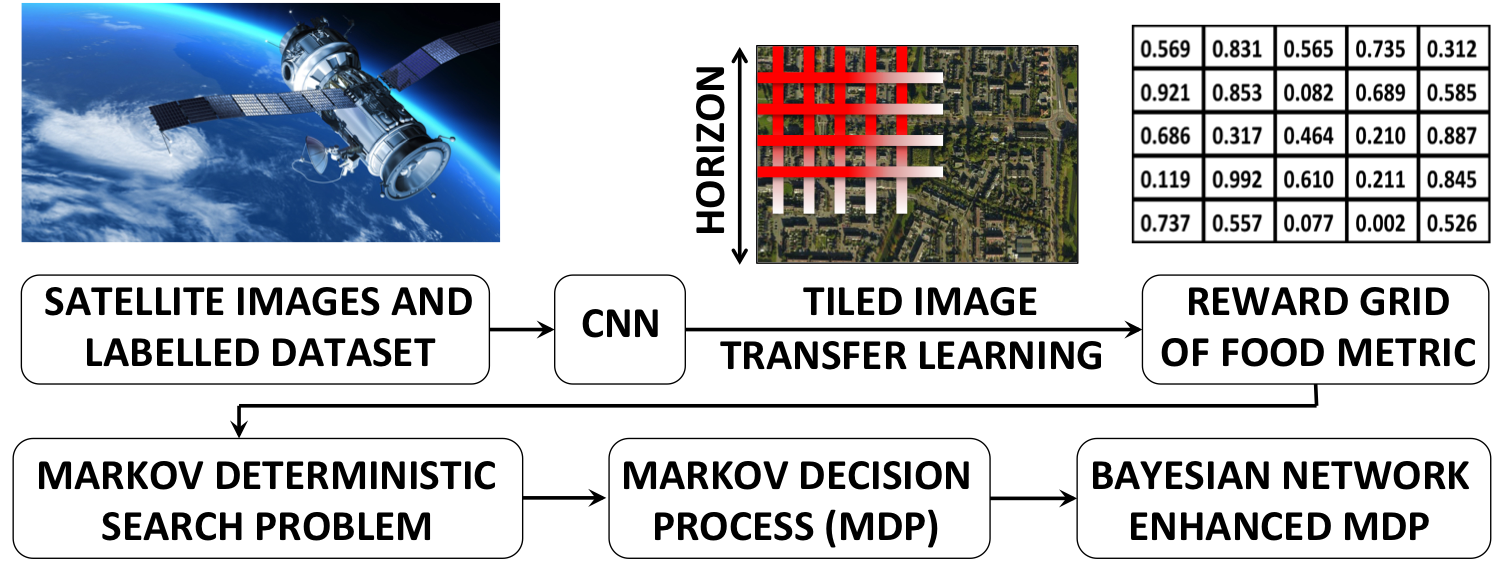}
\caption{A schematic describing the methodology for training CNN model which acts as an input for the path search algorithms}
\label{Meth}
\end{figure}
We can then tile the large 400 x 400 image into $33^2$ 12 x 12 tiles, which are then upsampled to 28 x 28 to match the resolution of the original training set. For each tile, we use our model to generate an associated food security metric which is treated as that tile's reward in a search problem.  Given this setup, we will sequentially tackle three linked sub-problems that each add a layer of complexity and uncertainty to solve the complete problem. These sequentially linked problems are described in the section titled ``Automated Satellite Tasking for Food Security Measurements".
A full diagram describing the setup of this problem can be found in Fig. \ref{Meth}.

\section{Predicting Food Security Metrics with CNNs}

\subsection{Approach}
Our first task is to effectively predict an FSM of interest directly from satellite data using a CNN.  There exist multiple potential mechanisms by which to approach this problem, and we perform this task with the aim both of generating a reward grid for satellite tasking as well as obtaining useful classification performance.  The first approach, which represents our baseline, entails using the pre-trained network of \cite{jean16} to extract features directly from the SustLab images and predict the relevant food security metric.  This network used a model originally trained on ImageNet to predict nighttime light levels from satellite imagery after transfer learning to fine-tune coefficients.  Features were extracted using this network and food security data was predicted from these features via ridge regression.  The output was transformed to a "classification" output by discretely binning the stunting percentage (percentage height-for-age Z-scores below two standard deviations of the reference population) in bins of $0$, $0-0.1$, $0.1-0.2$, $0.2-0.3$, $0.3-0.4$, and $\ge 0.4$, which yielded a reasonably Gaussian distribution of the training data to the different bins.  A "classification accuracy" could then be easily determined by comparing the predicted bin for each example to the bin in which its ground truth value fell.  Using this metric, the baseline attains a classification accuracy of 39 \%.

We attempt several different approaches to improving this baseline prediction.  First, we hypothesize that using a network pretrained on aerial imagery would yield features more useful in predicting food security metrics from satellite imagery than a network originally trained on ImageNet, which describes optical images of a variety of different objects with features very different than those observed in satellite images.  We therefore utilize the publicly available DeepSat-6 dataset of \cite{basu15} as an initial dataset, which consists of 405,000 28 x 28 x 4 images (at 1m resolution) labeled with one of six ground cover classifications: barren, grassland, trees, road, building, and water.  In order to enable transfer learning, it is optimal that the resolution of the two datasets match -- thus, we downsample the 1m DeepSat imagery to an effective resolution of 2.4m to match that of the SustLab data.  

Once good classification performance on this DeepSat dataset was observed, we performed a tiling operation very similar to that of \cite{basu15} on the SustLab data, where we resized each 400 x 400 image to 420 x 420 and then extracted 1089 12 x 12 tiles covering the full field of view, which were then upsampled to 28 x 28 x 3 to align with the dimensions of the DeepSat data.  Note that because the SustLab data did not include a fourth near-IR channel (only RGB), the near-IR channel was dropped from this analysis. This tiling procedure results in a dataset in excess of 500,000 image tiles, where each tile is labeled with the discrete bin commensurate with the stunting percentage value of the full image from which it was taken.  This tiled dataset can be used in three different ways: 1) direct learning, where we attempt to learn the image label directly from the tiles, 2) transfer initialization learning, where we perform a similar task, but initialize the network weights with those learned from training on the DeepSat classification task, or 3) transfer learning by directly using the DeepSat-trained model to extract features from SustLab tiles, then using ridge regression to compute a prediction of stunting percentage from these features.  

While it is possible to treat this as a regression problem, we have chosen to treat it as a classification problem due to the following considerations.  First, the data from which FSMs are drawn is survey data taken at a number of households within a given geographic area, the latitude and longitude of which are then correlated with those of the corresponding satellite image.  This data is fundamentally noisy in nature, and it is less important that an exact value be recovered than that the general state of food security in these areas be accurately reflected.  Thus, a classification framework seems appropriate, since binning the continuous results into discrete classes eliminates some of this noise.  Second, the $r^2$ metric characteristic of a regression heavily penalizes large single errors, whereas in this task, it is far more important to accurately classify an areas as "food secure" or "food insecure" than to be exactly correct about its stunting percentage value, and the differential practical cost of large errors and small errors in stunting percentage for single observations is negligible.

% nightlight baseline
% extracted 32 landcover features/ridge
% extracted 256 landcover features/ridge
% extracted 1024 landcover features/ridge
% transfer initialization + direct learning
% direct learning 

\subsection{Infrastructure}
To enable the studies described above, we created a custom codebase from scratch, which handles all aspects of the above approach including data extract-transform-load (ETL), CNN training/testing, and ridge regression/classification computation.  Note that all code can be found in the repository 

\subsubsection{ETL}
Data for the food security task is drawn from satellite images labeled with FSMs drawn from United Nations Demographic and Health Surveys (DHS).  Data is loaded from a set of .npy files describing the food security metrics of a target set of household surveys, and the location (in terms of latitude and longitude) is extracted from each location.  These metrics are applied to a square of 0.09 in lat-lon coordinate length centered on each reported survey location.  Next, for each image in the dataset for Nigeria, which we focus on here, we assign it to the cluster in which its location is contained.  All of this is accomplished in the \texttt{create\_clust\_dict2.py} file in 
\texttt{./test} in the code directory.  Next, in \texttt{./test/create\_test\_train\_set.py} a number of different options can be specified for creating test/train sets.  In our case, we tile one image from each survey cluster into 12 x 12 squares (afterwards upsampled to 28 x 28) and populate the train/test sets with an 80/20 split of tiles from each image.  

\subsubsection{Training and Monitoring}
\begin{figure}
\centering
\includegraphics[width=0.5\textwidth]{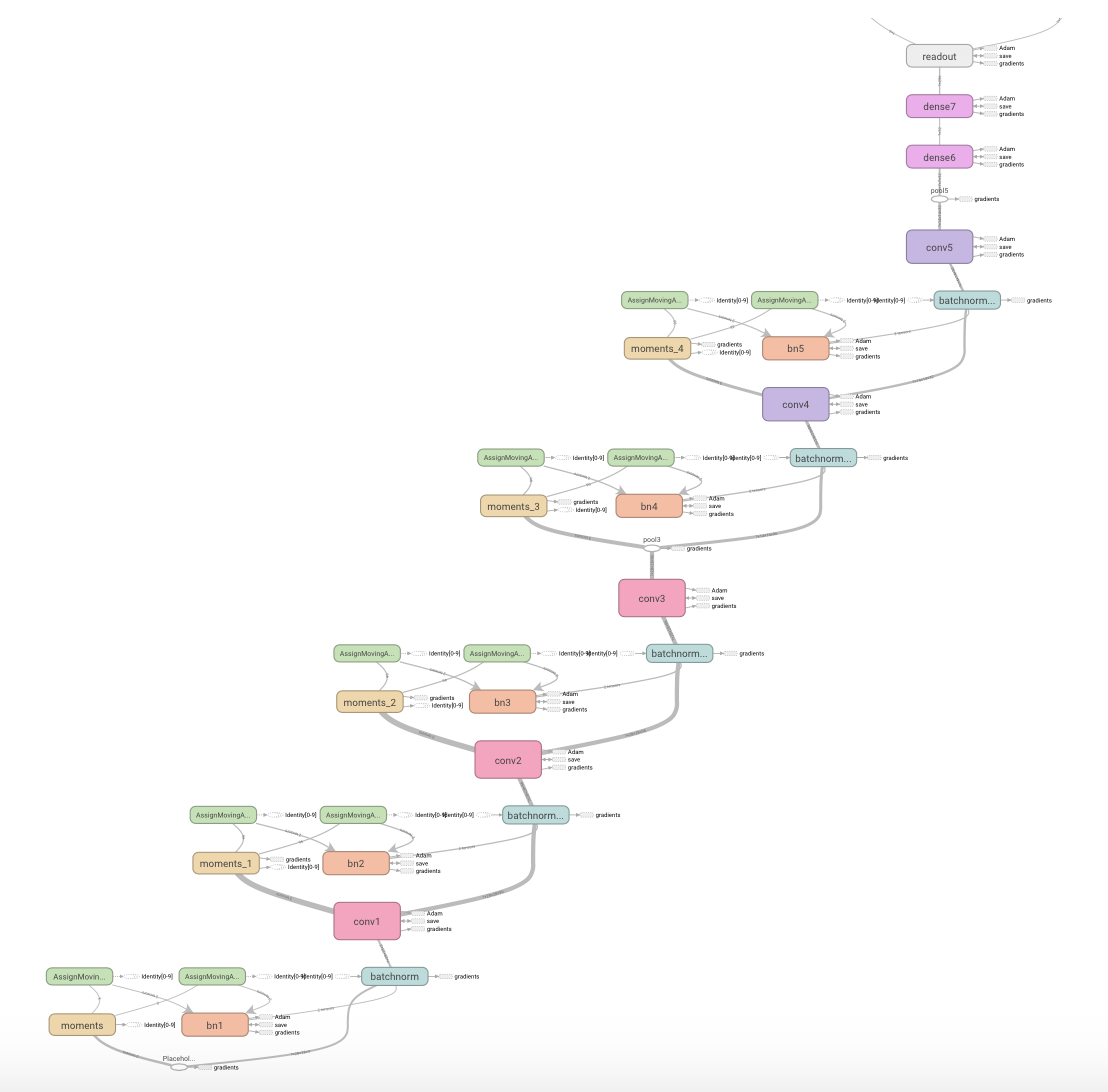}
\caption{Computational graph for VGG-style network.}
\label{compgraph}
\end{figure}

Utilities to create, train, and monitor the network can be found in the "\texttt{./utils}" directory, which contains several thousand lines of code.  For example, "\texttt{layer\_utils.py}" contains various layer types (i.e. general conv2d, batch normalization, maxpool), and several initialization functions (Xavier, He) incorporated from recent literature.  Model testing and training scripts can be found in "\texttt{./transfer\_learning}," which utilize the utilities in a modular fashion.  Further, Google TensorBoard functionality is integrated into the underlying training routines in a way that enables direct visualization of training curves, gradient flows, network activations, and the computational graph.  A computational graph for the VGG-style network used here for DeepSat classification can be found in Fig. \ref{compgraph}.  Further, completed notebooks for training, monitoring, and performance of ridge regression can be found in the "\texttt{./notebooks}" folder in the provided code repository.

\subsection{Results and Discussion}
\subsubsection{DeepSat Land Cover Classification Task}

Before performing the FSM prediction task, we first want to ensure that we have a viable network for land cover classification from which to perform transfer learning.  To accomplish this task, we have implemented a neural net architecture in the style of the classic VGGNet in TensorFlow, where the majority of layers are either 3 x 3 convolutional layers or 2 x 2 max pooling layers \cite{zisserman14}.  Using a GPU-powered environment allows us to substantially decrease training time (by over an order of magnitude) using an NVIDIA Titan X GPU as opposed to the CPU environment to which we were previously confined.  Our first pass at training a viable network was performed on the Deepsat image set, with an architecture as described in Table \ref{nettab}:

\begin{table}[h] \caption{Layout of VGG-Style Net Implementation.}
\label{nettab}
\centering
\begin{tabular}{|c|c|c|}
\hline
Layer & Type & Parameters \\ \hline
0 & Input & 4 x 28 x 28 \\ \hline
1 & Conv-Relu-Lrn & 96 x 3 x 3 \\ \hline
2 & Conv-Relu-Lrn & 96 x 3 x 3 \\ \hline
3 & Conv-Relu-Lrn & 96 x 3 x 3 \\ \hline
4 & Maxpool (2 x 2) & -  \\ \hline
5 & Conv-Relu-Lrn & 32 x 3 x 3 \\ \hline
6 & Conv-Relu-Lrn & 32 x 3 x 3 \\ \hline
7 & Conv-Relu-Lrn & 32 x 3 x 3 \\ \hline
8 & Maxpool (2 x 2) & -  \\ \hline
9 &FC & 32 \\ \hline
10 &Dropout & 0.9 \\ \hline
11 & Output & 6 \\ \hline
\end{tabular}
\end{table}
Note that in Table \ref{nettab}, "Lrn" refers to local response normalization, a procedure that normalizes pixels over a local spatial domain -- each layer was also batch normalized to optimize gradient propagation.

\begin{figure}
\centering
\includegraphics[width=0.5\textwidth]{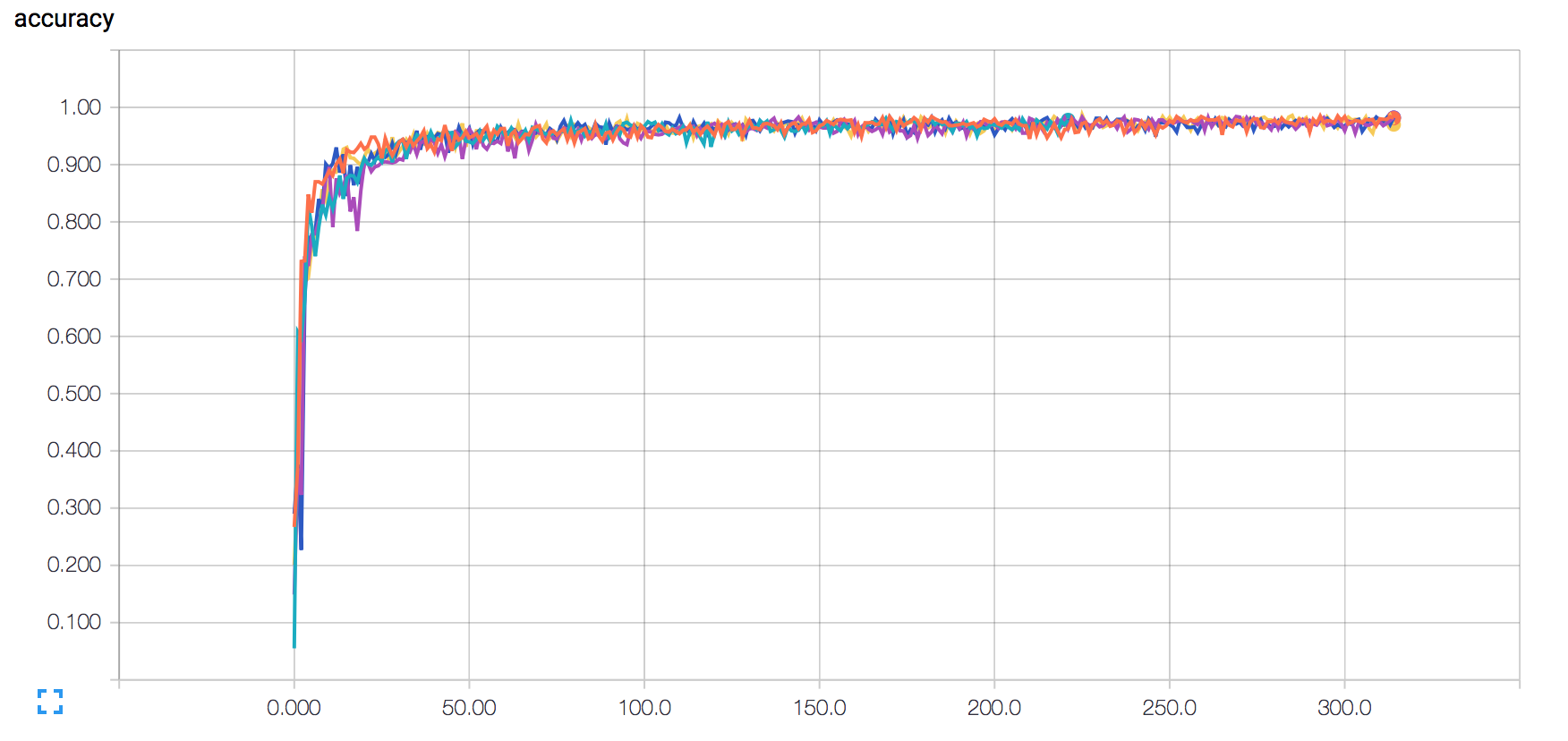}
\caption{Training curve for DeepSat classification task.}
\label{train}
\end{figure}

As shown in Fig. \ref{train}, training this network for one run over the full dataset yields a validation accuracy of 95\% and testing accuracy of 87 \% -- note that these are comparable to performance observed in \cite{basu15}, and additional training facilitated by the GPU environment should yield improved model performance.  We work with 80/20 test-train split with a validation set of 10,000 images.  We performed a hyperparameter search to ensure effective training, and we use local response normalization, dropout, and L2 regularization to avoid overfitting.  Our best parameters were 0.00071 for learning rate, 0.001 for the L2 regularization coefficient, and 0.1 for the dropout probability.  The confusion matrix for this network can be found below in Fig. \ref{CM}.

\begin{figure}
\centering
\includegraphics[width=0.45\textwidth]{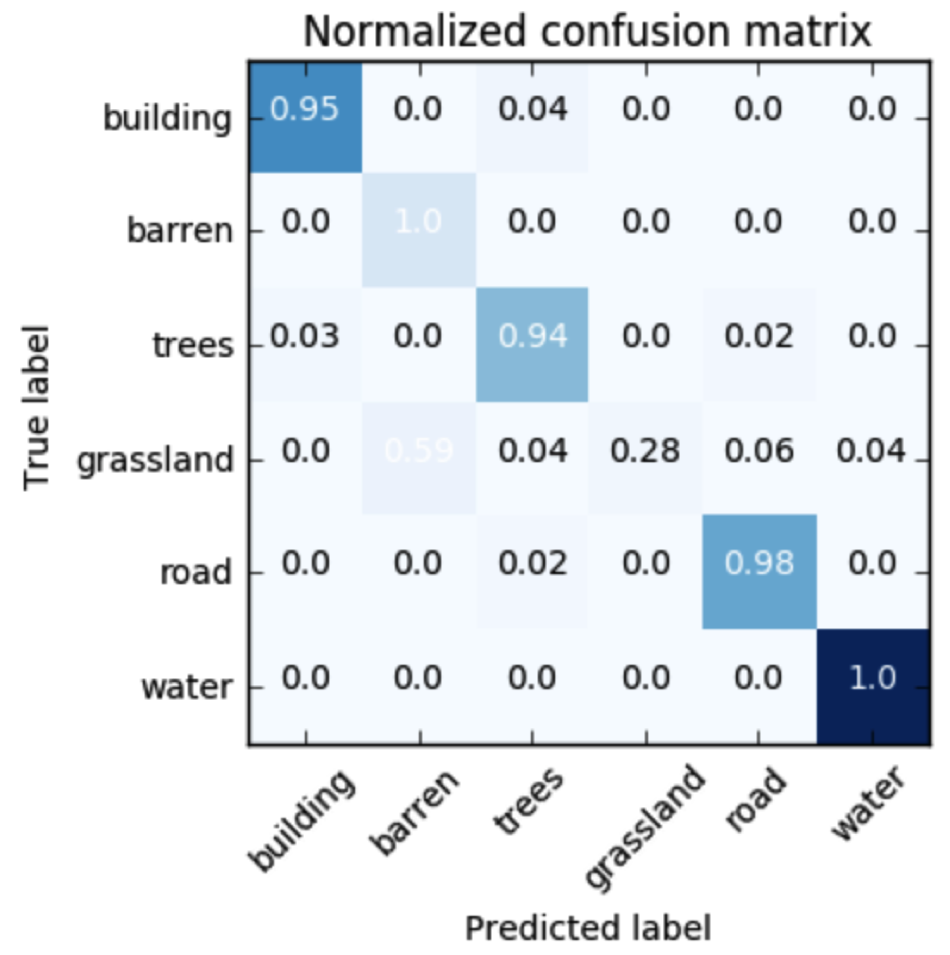}
\caption{Confusion matrix for the prediction of land cover classes}
\label{CM}
\end{figure}

	The high level of performance observed above may indicate that the network analyzes these images in a very similar way to how a human reader would -- for instance, the confusion matrix indicates that the most commonly confused classes are grassland and barren, which many humans would also have trouble differentiating.  These accuracy results in and of themselves are worth noting in comparison to the work of \cite{basu15}, who use labor-intensive hand-designed features and do not outperform this deep architecture by a significant amount.    

\subsubsection{Food Security Prediction: Stunting Percentage}
Now that we have attained useful results for the land cover classification problem, we can use this trained network to extract features from SustLab imagery that would be relevant for FSM prediction.  We use the exact same network architecture specified above in order to enable particularly straightforward transfer learning.   Table \ref{acctab} contains the results of several methods for FSM prediction, with stunting percentage as the target FSM.  While only test accuracies are reported here, we observed validation accuracies very similar to training accuracies, indicating that the network is not overfitting on our data.

% <Regression>
% <Classification>
% <Direct Learning, Transfer Learning>

\begin{table*}[h!] \caption{CNN FSM Prediction Results.}
\label{acctab}
\centering
\begin{tabular}{|c|c|c|c|}
\hline
Model & Output Layer Feature Number & Tile or Full Image & Test Accuracy  \\ \hline
Nightlight Features & 4092 & Full Image & 39\% \\ \hline
Image-Mean DeepSat Features + RR & 32& Full Image & 31.1 \%   \\ \hline
Image-Mean DeepSat Features + RR & 256  & Full Image & 30.8 \% \\ \hline
Direct CNN Learning & 32 & Tile &  43.8 \% \\ \hline
Direct CNN Learning & 256 & Tile &  45.0 \% \\ \hline
Direct CNN Learning & 1024 & Tile & 46.5 \% \\ \hline
Direct Learning with Transfer Initialization & 256 & Tile & 46.4 \% \\ \hline

\end{tabular}
\end{table*}

We can now elaborate upon several of the prediction procedures described in Table \ref{acctab}. One method, referred to as "Image-Mean DeepSat Features + RR," utilized the average of the features extracted from the SustLab tiles for each full image and used ridge regression to predict the full image stunting percentage.  This gives accuracy nearly 10 \% worse than the baseline classification.  We suspect that averaging may cause some features captured by activations to be lost, though there may be alternative strategies.  The "Direct CNN Learning" models directly train the network on SustLab image tiles labeled with the stunting percentage of their parent image.  As shown here, the number of features does not play a particularly important role in the test accuracy of the classification output, which is generally around 45 \%.  These results imply that if given a single 28m x 28m tile from a given image is supplied, we can predict the stunting percentage of that image correctly to within 10 \% of the actual value with 50 \% probability, which is surprising.  Furthermore, we are able to outperform the nightlight baseline by over 5 \% just based on these tile scale predictions, which implies that some features relevant to stunting are on the order of less than 28 meters.  In the future, it would be helpful to look at layer activations to determine the exact nature of the features that are providing this predictive capacity.  We also attempted transfer learning in a different manner, by initializing the network with weights learned for DeepSat, then continuing to train directly on the SustLab images and associated stunting percentage.  This procedure slightly outperformed the pure Direct CNN Learning approach with a similar number of output layer features (256), but did not provide substantial improvement.

\section{Automated Satellite Tasking for Food Security Measurements}\label{algos}
\subsection{Images to Reward Grids: Methodology and Algorithms}
The prediction pipeline obtained from the analysis above produces a food security metric (FSM) given an image, which becomes an input for an automated satellite tasking algorithm. The FSM we use is the stunting percentage, which is normalized between 0 and 1 such that the numbers closer to 0 denote food insecure regions whereas numbers closer to 1 denote regions with higher food security.

\begin{figure*}
\centering
\includegraphics[width=0.75\textwidth]{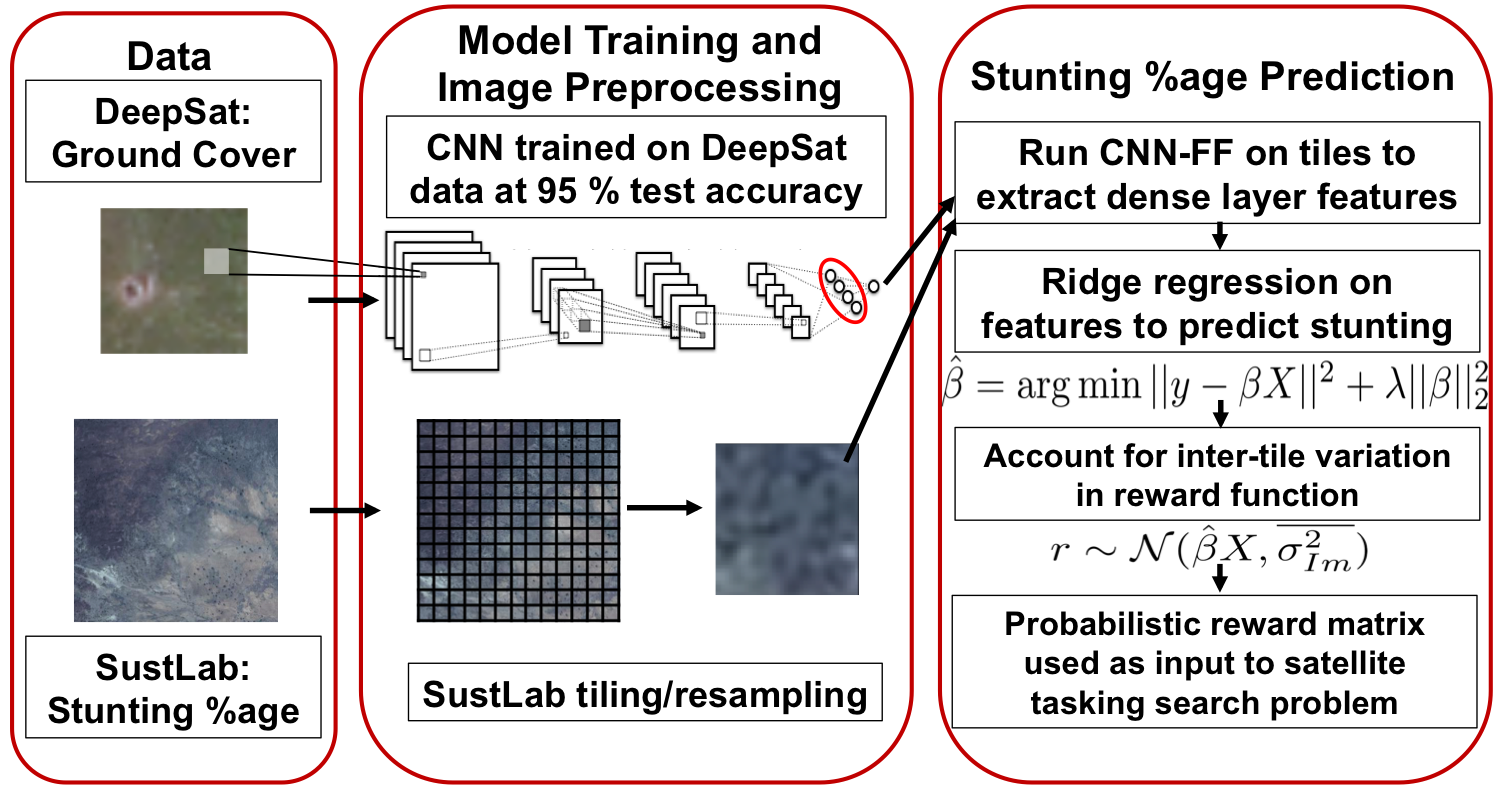}
\caption{A schematic for the path from an image to generating an FSM}
\label{res1}
\end{figure*}
\begin{figure}
\includegraphics[width=0.5\textwidth]{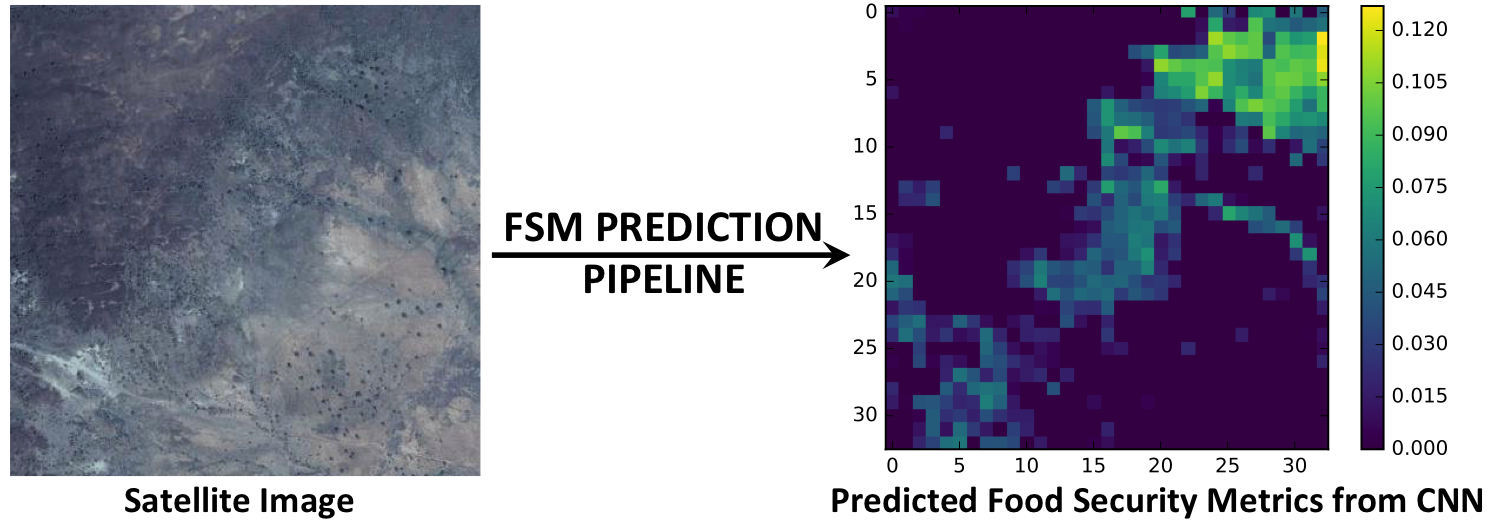}
\caption{Visualization of the high resolution FSM predicted by our model}
\label{vis}
\end{figure}

Figure \ref{Meth} shows the project pipeline. In the prior sections, we have discussed the first stage in this methodology where we train a model on the labeled images from DeepSat \cite{basu15} and SustLab \cite{jean16}.

This excercise helps us train a model that can predict the FSMs over various sub-tiles of a larger image. To give an analogy, even though the Bay Area is food secure, sub-regions within the Bay Area are not homogenous (i.e. Palo Alto versus East Palo Alto). In other words, transfer learning helps us go from a low resolution prediction to a high resolution prediction. The process of extracting a reward grid from the tiled image is the quantitative description of this idea. The reward grid is produced by tiling the large image into small regions (making sure that the resolution of the training dataset for the model matches the resolution of the tiles) and then using the trained model to predict the FSM at each tile.  Once this matrix has been created, we can look at the satellite tasking problem as a set of three sequentially linked sub-problems that each add a layer of complexity and uncertainty to solve the complete problem. After pre-processing the SustLab images, we end up with a 33 x 33 grid of 28 x 28 pixel images.\\  

\begin{figure}
\includegraphics[width=0.5\textwidth]{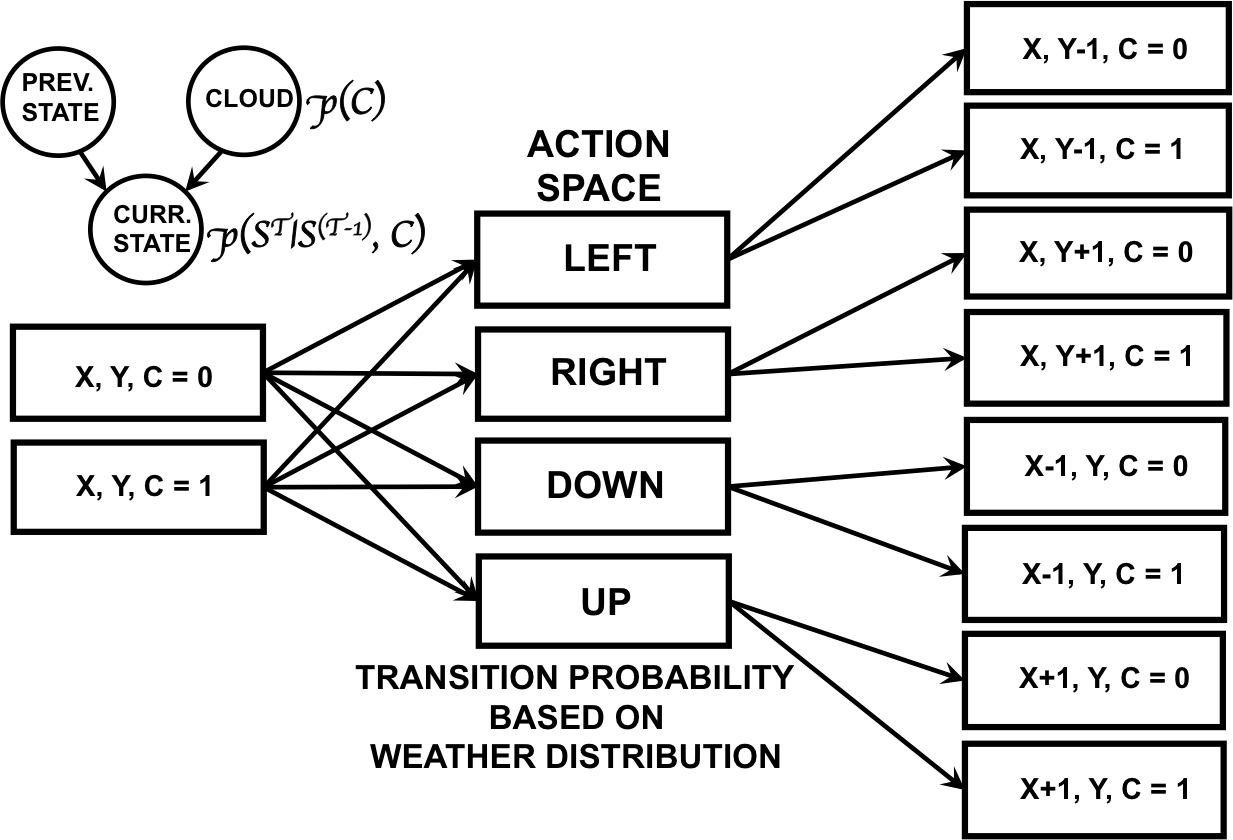}
\caption{The Bayesian Network driven Markov Decision Process. The Bayesian Network completely specifies the weather distribution which in turn completely specifies the MDP}
\label{BN}
\end{figure} 

\textbf{Sub-Problem 1: Finite Horizon Deterministic Approach:} Given our trained CNN H, we can use the constant time prediction property of neural networks to traverse through each tile of the large image and create a 33 x 33 grid of predicted food security metrics in $\mathcal{O}(33^2)$ time. On this grid, we can start at a randomly chosen location and set up a Markovian search problem to find the tile with the lowest food metric (corresponding to the poorest region in our map) in the the least number of tile-to-tile moves (we choose the number of moves as the metric of cost and define an optimal path as the least cost path, acknowledging that this path is not guaranteed to be unique). We use Uniform Cost Search for this sub-problem.\\

\textbf{Sub-Problem 2: Finite Horizon Stochastic Approach: Aleatoric Uncertainty:} In the previous model of the problem, we assumed that there are bounded and finite horizons to our search space. We also spent time traversing each tile to calculate the food security metric to set up the Markovian search problem. Inherent in the prediction task is an error incurred from the model having a finite accuracy. As discussed above, a ridge regression is performed on the features produced by the CNN when the satellite image is the input. Therefore, the prediction of the FSM that is obtained is the expected value of the FSM with variance characterized by the coefficient of determination of the regression fit. Recall that the coefficient of determination is the deviation from unity of the ratio of the residual sum of squares and the total sum of squares. In a general form, $R^2$ can be seen to be related to the fraction of unexplained variance (FUV), since the ratio compares the unexplained variance (variance of the model's errors) with the total variance (of the data). To get an estimate of this prediction variance, we use the following procedure:\\
\begin{algorithmic}
\State Input: Images tagged with FSMs
\For{Image i = 1, 2, 3, $\ldots$}  
    \For{On Image i, Tile t = 1, 2, 3, $\ldots$} 
    	\State {Get FSM Prediction on tile t. Call it $P_{it}$}
    \EndFor
    \State {$Var_{i}$ = Variance of $P_{it}$ over all t}
\EndFor
\State {Global Variance = $\sigma^2$ = Average of $Var_{i}$ over i}
\end{algorithmic}~\\
This variance characterizes up to the second moment of the background probabilistic distribution from which the reward grid is predicted. Sampling from this distribution and using the Monte Carlo approach to calculate the relevant statistics allows us to account for this uncertainty and calculate a prediction error margin. This margin is calculated for the predicted optimal path. The schematic in figure \ref{res1} and the following procedure outlines our approach.\\
\begin{algorithmic}
\State {Input: Expected Reward Grid ($\mu$), FSM Model Prediction Variance ($\sigma^2$)}
\State {Each tile on reward grid has added parameter \textbf{onPath}}
\State {Create a pathProbabilityMatrix of size equal to the size of the reward grid. Initialize this matrix to zeros}
\For{Monte Carlo Realization r = 1, 2, 3, $\ldots$, MCIters}
	\For{On Reward Grid, Tile t = 1, 2, 3, $\ldots$}  
    	\State {Reward Value at tile t sampled from $\mathcal{N}$($\mu_{t}$, $\sigma^2$)}
    	\State {Perform UCS on this realization of reward grid}
        \State {Assign onPath = 1 to tiles that are on optimal path}
        \State {Assign onPath = 0 to tiles that are not on path}
	\EndFor
    pathProbabilityMatrix += onPathMatrix 
\EndFor
\State {pathProbabilityMatrix /= MCIters} 
\end{algorithmic}~\\
The path probability matrix assigns a probability of the optimal path passing through a given tile. This probability serves as a metric for a two-pronged validation of the proposed model. One, a sharper path with less variance implies that the trained model works well and is sufficient for the given geographic area and builds confidence in our prediction of FSMs. Two, a large variance indicates a mismatch between the dataset on which training is conducted versus the dataset on which prediction is attempted and signals that either increasing the training database or adding more geographically relevant data into the database would be advisable.\\

\begin{figure*}
\includegraphics[width=\textwidth]{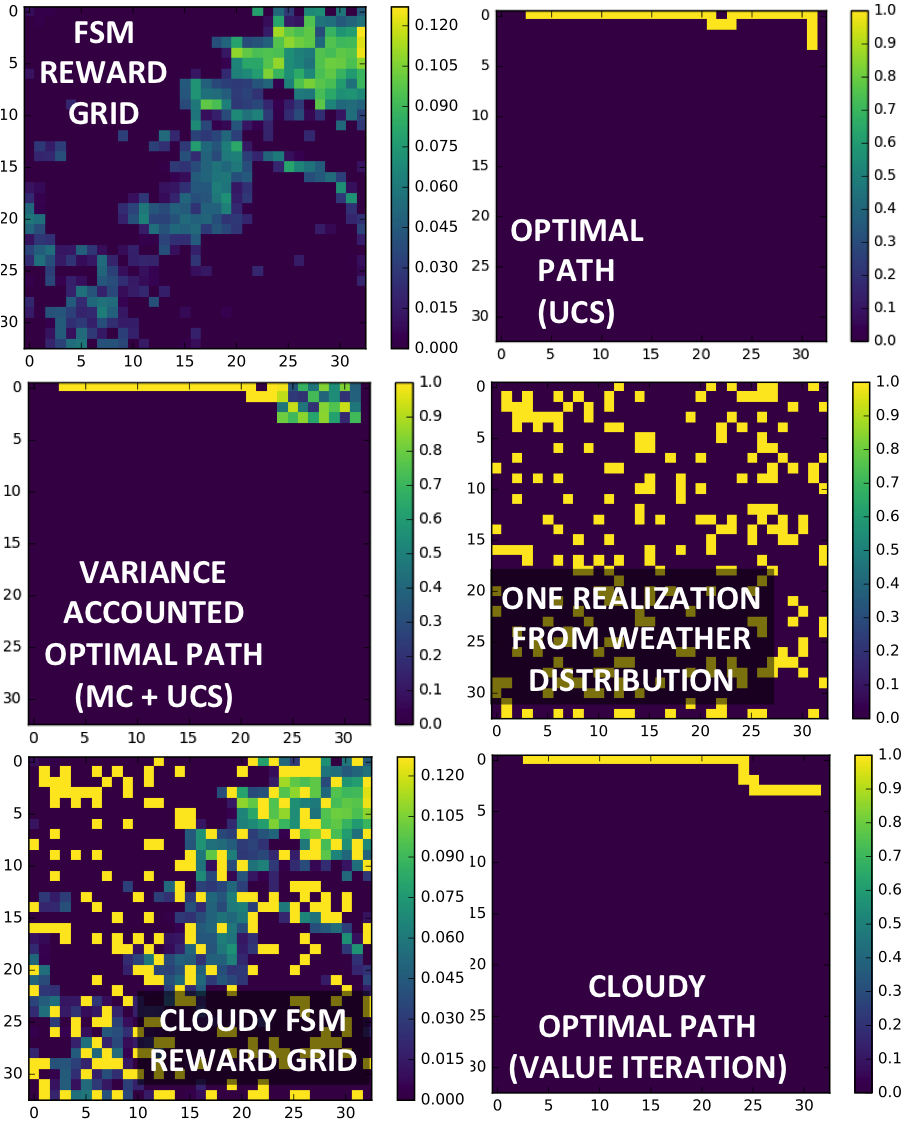}
\caption{The results derived after an image has gone through the entire pipeline}
\label{res}
\end{figure*}

\textbf{Sub-Problem 3: Finite Horizon Stochastic Approach: Epistemic Uncertainty:} When satellite images are used, the natural question that comes up is ``What happens when the satellite view is occluded due to clouds?" To address this, we propose a Markovian cloud cover model with Bernoulli probabilities for cloudy tiles. This model is completely specified by the Bayesian network in Fig. \ref{BN}. Thus, the future step in the path search depends on the cloud cover ($p(C)$) that exists in the current state and the probability of transitioning ($p(C^{t}|C^{t-1})$) to a tile with/without cloud cover given the current state of the cloud cover. An additional binary variable is added to our minimal description of the state to account for this. Note that both these probabilities (which has a degree of freedom 3 for the 6 required probabilities) can be estimated from historical data using a frequentist estimation strategy like maximum likelihood. A more physical weather model would incorporate spatial correlations and temporal statistics which can be modeled using a deeper Bayesian network and a higher dimensional state-space. Note that the runtimes of all the path search algorithms used scale exponentially in the size of the state-space. For the purposes of this project and a proof-of-concept, our limited validity model suffices. Once both $p(C)$ and $p(C^{t}|C^{t-1})$ are prescribed, the epistemic uncertainty is incorporated into the problem specification and transforms the deterministic search problem into a Markov Decision Process. The state subgraph of the full MDP search graph is schematically shown in Fig. \ref{BN}. On this MDP, value iteration is performed so as to obtain the optimal movement policy given a state. It is instructive to note that once this policy is computed, we do not need to re-compute the optimal policy since prescribing the probability space of the MDP completely determines the optimal policy. This is the desired and required behavior for an algorithm that needs to be deployed on light hardwares - typical of satellites. We perform this value iteration process on our MDP with $p(C = 1) = 0.2$, $p(C^{t} = 1|C^{t-1} = 0) = 0.5$ and $p(C^{t} = 1|C^{t-1} = 1) = 0.5$. The figure on the right in the middle row of Fig. \ref{res} shows one particular realization of the cloud cover. The figure on the left in the bottom of Fig. \ref{res} shows the occluded view seen from a satellite due to the cloud cover in contrast to the clean view (top left of Fig. \ref{res}) that would otherwise be expected to be seen on a clear day. Value Iteration, in contrast to UCS tries to maximize the reward. The reward matrix described above is scaled affinely and linearly so that all values lie between 0 and 1 and the values close to 1 signify lower food security. The goal is once again to reach the most food insecure point while traversing food insecure regions such the path is globally optimal in the sense that the locations traversed pass through the most food insecure regions within the horizon. In order to make sure that the path search avoids clouds, a zero value (high food security) is assigned to the locations where clouds exist. Shown on the bottom right in Fig. \ref{res} is the optimal path found when the realized cloud cover occludes the view. It is seen that the optimal policy learns to meander around the clouds. The optimal path in this case is also significantly different than the optimal path obtained from UCS.        

\section{Conclusions and Future Work}

In the end, we have accomplished several goals with this work.  First, we have shown that CNNs can accurately predict land cover classification on the DeepSat dataset with up to 87 \% accuracy, which is on par with state-of-the-art results from other methods.  Second, we have shown that specification of a tile-scale CNN model either with direct training on tiled SustLab images or via transfer initialization from a network trained on DeepSat data yields accuracy of over 45 \% on the test set, which outperforms the baseline of 39 \% generated by the current state-of-the-art.  The fact that this tile-scale CNN sourced from training on satellite-relevant datasets performs better than features extracted on the whole images from nightlight data implies that direct training on the metric of interest is indeed a useful contributor to improved accuracy.  In the future, there are a variety of directions that the machine learning side of this work could be taken, including iterating over network architectures (GoogLeNet, ResNet, etc.) and obtaining additional data to allow for direct training on the larger-scale image. Use of either GoogLeNet or ResNet seems particularly alluring given their ability to incorporate (and simultaneously output) features at multiple different scales while incorporating skip connections, which may be particularly helpful in ensuring that lower-level features are emphasized in the classification layer.

On the satellite tasking portion of our project, we have shown that tasking under uncertainty can be handled both by UCS with a CNN-based probabilistic cost function and MDPs driven by Bayesian networks.  Future work on this front should include incorporating physical, non-Markovian weather models to more accurately simulate actual satellite conditions and identification of high-resolution satellite assets that could benefit from these types of algorithms in practice.

\section{ Acknowledgments}
Prof. Stefano Ermon, Prof. Marshall Burke, and Stanford Sustainability and Artificial Intelligence Lab

\clearpage
\bibliographystyle{aaai}
\bibliography{bibliography}

\end{document}